\documentclass[conference]{IEEEtran}
\IEEEoverridecommandlockouts
\usepackage{cite}
\usepackage{amsmath,amssymb,amsfonts}
\usepackage{microtype}
\usepackage{algorithm}
\usepackage{algorithmic}
\usepackage{graphicx}
\usepackage{booktabs}
\usepackage{multirow}
\usepackage{threeparttable}
\usepackage{tabularx}
\usepackage{url}
\usepackage{textcomp}
\usepackage{xcolor}
\usepackage{placeins} 
\usepackage{tikz}
\usepackage{listings}
\usetikzlibrary{arrows.meta, positioning}

\definecolor{kernelbg}{RGB}{245,245,245}
\definecolor{designbg}{RGB}{227,242,253}

\newcommand{\DiffHLS}{\textsc{DiffHLS}}
\usepackage[hidelinks]{hyperref}
\def\BibTeX{{\rm B\kern-.05em{\sc i\kern-.025em b}\kern-.08em
    T\kern-.1667em\lower.7ex\hbox{E}\kern-.125emX}}
\begin{document}

\title{\DiffHLS: \underline{Diff}erential Learning for \underline{H}igh-\underline{L}evel \underline{S}ynthesis QoR Prediction with GNNs and LLM Code Embeddings}

\author{\IEEEauthorblockN{Zedong Peng}
\IEEEauthorblockA{\textit{Shanghai Jiao Tong University}\\
Shanghai, China \\
zedongpeng@sjtu.edu.cn}
\and
\IEEEauthorblockN{Zeju Li}
\IEEEauthorblockA{\textit{The Chinese University of Hong Kong}\\
Hong Kong, China \\
lizeju@link.cuhk.edu.hk}
\and
\IEEEauthorblockN{Qiang Xu}
\IEEEauthorblockA{\textit{The Chinese University of Hong Kong}\\
Hong Kong, China \\
qxu@cse.cuhk.edu.hk}
\and
\IEEEauthorblockN{Jieru Zhao}
\IEEEauthorblockA{\textit{Shanghai Jiao Tong University}\\
Shanghai, China \\
zhao-jieru@sjtu.edu.cn}
}

\maketitle

\begin{abstract}
High-Level Synthesis (HLS) compiles C/C++ into RTL, but exploring pragma-driven optimization choices remains expensive because each design point requires time-consuming synthesis. We propose \textbf{\DiffHLS}, a differential learning framework for HLS Quality-of-Result (QoR) prediction that learns from kernel--design pairs: a kernel baseline and a pragma-inserted design variant. \DiffHLS~encodes kernel and design intermediate-representation graphs with dedicated graph neural network (GNN) branches, and augments the delta pathway with code embeddings from a pretrained code large language model (LLM). Instead of regressing absolute targets directly, we jointly predict the kernel baseline and the design-induced delta, and compose them to obtain the design prediction. On PolyBench, \DiffHLS~attains lower average MAPE than GNN baselines under four GNN backbones, and LLM code embeddings consistently improve over a GNN-only ablation. We further validate scalability on the ForgeHLS dataset.
\end{abstract}


\section{Introduction}
HLS compiles C/C++ into RTL and is widely used to build FPGA accelerators. In practice, designers explore pragma choices such as pipelining and unrolling to trade off performance and resource usage, but each design point can require a costly HLS run.

QoR prediction estimates post-HLS design metrics to accelerate design space exploration (DSE); typical targets span resource utilization metrics such as DSP, FF, and LUT, and timing metrics such as the critical-path delay (CP). GNN-based predictors over IR graphs are effective~\cite{wu2022high,gao2024hierarchical}, but they must handle large inter-kernel target variance and may miss pragma-aware semantic cues that remain explicit in source code.

This paper introduces \textbf{\DiffHLS}, a differential QoR predictor that learns from kernel--design pairs. We treat the original kernel as a baseline and a pragma-inserted implementation as a design variant, and train the model to predict both the kernel baseline and the design-induced delta. We further incorporate frozen LLM code embeddings to complement IR graph signals. The key design rationale behind \DiffHLS~rests on two observations:

1) Predicting changes is more precise than predicting absolutes.
In pragma-driven datasets, each kernel induces a cluster of design points that share a large kernel-dependent baseline but differ due to pragma choices. Directly regressing the design-level target forces the model to simultaneously explain the large baseline differences across kernels and the smaller, pragma-driven variations within each kernel. By decomposing the prediction into a kernel baseline and a design-induced delta, we reduce target variance within each cluster, let the model focus on pragma effects, and gain a natural diagnostic: errors can be attributed to either baseline estimation or delta modeling.

2) Graphs and code carry different but complementary information.
In HLS optimization, pragma directives and code-level patterns such as loop structure and memory-access style strongly influence resource usage. IR graphs capture control and data dependencies across the program, while source code retains explicit pragma annotations and local patterns that may be weakened during IR lowering. We therefore combine graph representations with pretrained LLM code embeddings: graphs model structural dependencies, and code embeddings provide pragma-aware semantic context.

Our contributions are:
\begin{itemize}
\item We propose \DiffHLS, a differential learning formulation for HLS QoR prediction that models each pragma-inserted design as a kernel baseline plus a learned delta.
\item We design a hybrid predictor that fuses CDFG-based GNN representations with pretrained LLM code embeddings to capture both IR-level structure and pragma-aware code semantics.
\item We evaluate on PolyBench with targeted ablations that quantify the contributions of differential learning, code embeddings, and component-level error sources, and further validate scalability on the larger ForgeHLS dataset.
\end{itemize}

\begin{figure*}[!t]
    \centering
    \includegraphics[width=0.75\textwidth]{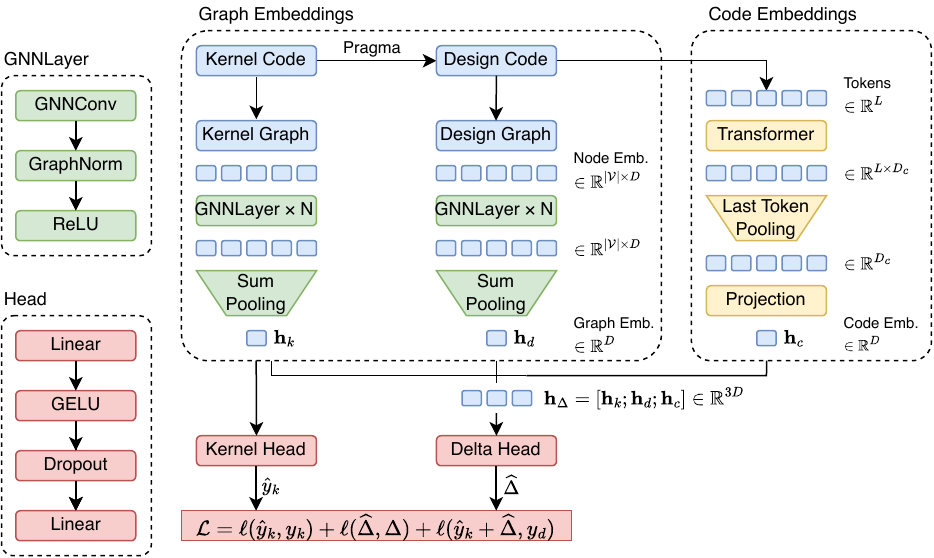}
    \vspace{-0.8em}
    \caption{Overview of \DiffHLS. We encode kernel and design IR graphs, inject LLM code embeddings into the delta pathway, and predict the kernel baseline and delta; finally the design prediction is obtained by composition.}
    \label{fig:overview}
    \vspace{-1em}
\end{figure*}

\begin{figure}[!t]
\centering
\begin{tikzpicture}[
    node distance=8mm,
    every node/.style={inner sep=6pt}
]

\node (kernel) [
    draw,
    rounded corners=3pt,
    fill=designbg,
    text width=0.9\columnwidth
] {
\vspace{-0.2em}
\begin{lstlisting}
void vector_add(int a[8], int b[8], int c[8]) {
    for (int i = 0; i < 8; i++) {
        c[i] = a[i] + b[i];
    }
}
\end{lstlisting}
};

\node (design) [
    draw,
    rounded corners=3pt,
    fill=designbg,
    text width=0.9\columnwidth,
    below=8mm of kernel
] {
\vspace{-0.2em}
\begin{lstlisting}
void vector_add(int a[8], int b[8], int c[8]) {
#pragma HLS ARRAY_PARTITION variable=a factor=2
#pragma HLS ARRAY_PARTITION variable=b factor=2
#pragma HLS ARRAY_PARTITION variable=c factor=8
    for (int i = 0; i < 8; i++) {
#pragma HLS PIPELINE OFF
#pragma HLS UNROLL factor=2
        c[i] = a[i] + b[i];
    }
}
\end{lstlisting}
};

\draw[-{Latex[length=2mm]}, thick]
(kernel.south) -- node[right, font=\footnotesize]
{Pragma insertion} (design.north);

\end{tikzpicture}

\caption{Example kernel code (top) and design code (bottom) illustrating HLS pragma insertion.}
\label{fig:code_examples}
\vspace{-1em}
\end{figure}
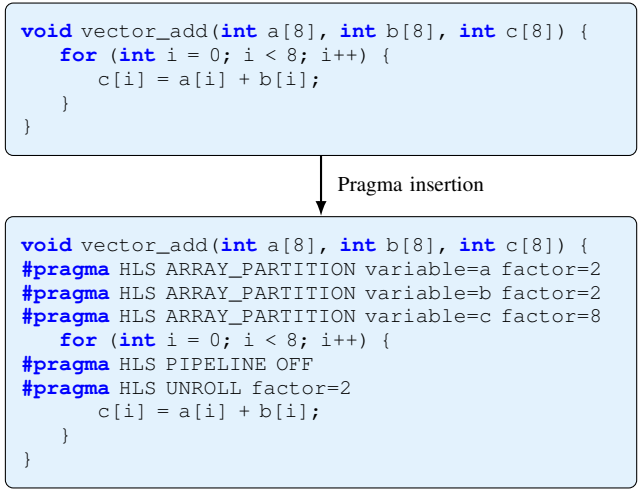

\section{Related Work}
\subsection{HLS QoR Prediction}
QoR prediction for HLS has been widely studied. 
Common benchmarks include PolyBench/C~\cite{Polybench}, DB4HLS~\cite{Db4hls}, and HLSyn~\cite{bai2023hlsyn}. Larger-scale datasets such as ForgeHLS~\cite{peng_forgehls_2025} and HLS4ML~\cite{hawks2025wa} expand diversity. 
Early work leverages hand-crafted features from HLS reports or compiler IR with classical regressors~\cite{zhong2017design,dai2018fast,zhao2017comba}.
Recent approaches represent programs as IR graphs and apply GNNs; representative examples include CDFG-based prediction~\cite{wu2022high}, hierarchical and heterogeneous extensions~\cite{gao2024hierarchical,sohrabizadeh2022automated,lin2022powergear,9045442}, post-route estimation~\cite{goswami2022predictingqor}, and GNN-based area and timing prediction~\cite{jamal2023graph}. Hierarchy-aware encoders target long-range dependencies from nested loops~\cite{sohrabizadeh2023robust,murphy2024balor}; related modeling has been studied for pre-HLS power prediction~\cite{lin2025hippo}. Cross-modal approaches fuse source-code tokens with IR graphs via transformers and GNNs~\cite{qin2024crossmodality}. Mixture-of-experts routing has been proposed for specialized prediction~\cite{li2025hierarchical}.
These methods typically regress absolute QoR targets per design, making it hard to capture the subtle effects of individual pragmas accurately. \DiffHLS~addresses this by decomposing prediction into a kernel baseline plus a pragma-induced delta.

\subsection{LLM Code Embeddings}
Pretrained code models provide useful representations for downstream code understanding; encoder-style models such as CodeBERT~\cite{feng2020codebert} and GraphCodeBERT~\cite{guo2020graphcodebert} learn semantic embeddings that can complement graph-structured IR features. Since many strong code LMs are decoder-only, recent work studies how to extract encoder-style representations from them~\cite{suganthan2025adapting}; other work explores using LM decoding directly for regression~\cite{song2025decoding}.
Code embeddings can complement IR graphs by retaining pragma directives and local semantic cues that may be weakened during IR lowering; this motivates lightweight fusion of pretrained code representations with graph encoders~\cite{feng2020codebert,guo2020graphcodebert}. \DiffHLS~incorporates a lightweight, frozen LLM code embedding as an auxiliary design feature, avoiding expensive LLM fine-tuning while improving sensitivity to pragma-driven changes.

\section{Method}
\subsection{Problem Formulation}
For each kernel $k$ and one of its pragma-inserted designs $d$, we predict a QoR target $y_d \in \mathbb{R}$, where the target is one of DSP, FF, LUT, or CP. We treat each design as a paired sample $(k,d)$ and decompose the design-level target into a kernel baseline $y_k$ and a design-induced delta:
\begin{equation}
    \Delta = y_d - y_k.
\label{eq:delta_def}
\end{equation}
This formulation reduces inter-kernel scale variance and centers learning on pragma effects.
Both $y_k$ and $y_d$ are obtained from post-HLS synthesis reports: $y_k$ represents the QoR metric of the original kernel code without pragma directives, while $y_d$ represents the corresponding metric of the pragma-inserted design code.

\begin{algorithm}[t]
    \caption{Training of \DiffHLS}
    \label{alg:diffhls}
    \begin{algorithmic}
    \STATE \textbf{Input:} paired samples $\{(\mathcal{G}_k,\mathcal{G}_d,y_k,y_d)\}$, code tokens $\mathbf{t}$
    \FOR{each minibatch}
    \STATE $\Delta \leftarrow y_d - y_k$
    \STATE $\mathbf{h}_k \leftarrow \text{GNNEnc}_k(\mathcal{G}_k)$; $\mathbf{h}_d \leftarrow \text{GNNEnc}_d(\mathcal{G}_d)$
    \IF{Code Emb.}
        \STATE $\mathbf{h}_c \leftarrow \text{CodeEnc}(\mathbf{t})$
        \STATE $\mathbf{h}_{\Delta} \leftarrow [\mathbf{h}_k;\mathbf{h}_d;\mathbf{h}_c]$
    \ELSE
        \STATE $\mathbf{h}_{\Delta} \leftarrow [\mathbf{h}_k;\mathbf{h}_d]$
    \ENDIF
    \STATE $\hat{y}_k \leftarrow f_k(\mathbf{h}_k)$; $\widehat{\Delta} \leftarrow f_{\Delta}(\mathbf{h}_{\Delta})$
    \STATE $\mathcal{L} \leftarrow \ell(\hat{y}_k,y_k)+\ell(\widehat{\Delta},\Delta)+\ell(\hat{y}_k+\widehat{\Delta},y_d)$
    \STATE Update parameters using $\nabla \mathcal{L}$
    \ENDFOR
    \end{algorithmic}
\end{algorithm}

\subsection{Graph Encoding}
We build IR graphs $\mathcal{G}_k$ and $\mathcal{G}_d$ for the kernel and design, respectively, following the CDFG (control-data flow graph) representation of Wu \emph{et al.}~\cite{wu2022high}. The graphs are extracted from the HLS toolchain's ADB intermediate representation, which captures both control flow and data dependencies in a unified graph structure.

We encode them with two parameter-disjoint graph encoders for the kernel and design branches, with the same architecture but independent parameters:
\begin{equation}
    \mathbf{h}_k = \text{GNNEnc}_k(\mathcal{G}_k), \qquad
    \mathbf{h}_d = \text{GNNEnc}_d(\mathcal{G}_d).
\label{eq:graph_enc}
\end{equation}
Here $\mathbf{h}_k,\mathbf{h}_d \in \mathbb{R}^{D}$. As illustrated in Fig.~\ref{fig:overview}, each encoder stacks $N$ GNN layers, where each layer consists of a graph convolution, followed by GraphNorm and a ReLU nonlinearity. A sum pooling readout aggregates node representations into a graph-level embedding.

We follow Wu \emph{et al.}~\cite{wu2022high} for node and edge attribute construction.

\subsection{LLM Code Embedding}
IR graphs capture compiler-level structure but may miss source-level patterns relevant to optimization. We therefore augment each design $d$ with a frozen code encoder $g(\cdot)$ from a pretrained code LLM. Let $\mathbf{t}$ denote the token sequence obtained by concatenating the design source files. We compute a fixed-length representation by extracting the last token's hidden state:
\begin{equation}
    \mathbf{z} = g(\mathbf{t})_{\text{last}}.
\label{eq:code_emb}
\end{equation}
Since the code embedding dimension $L$ may differ from the GNN hidden size $D$, we use a lightweight adapter $a:\mathbb{R}^{L}\!\to\!\mathbb{R}^{D}$:
\begin{equation}
    \mathbf{h}_c = a(\mathbf{z}).
\end{equation}
Thus $\mathbf{h}_c \in \mathbb{R}^{D}$.
We inject LLM code embeddings only into the delta pathway via the delta head, to preserve the interpretation of $y_k$ as a GNN-only kernel baseline.

For each design $d$, we use its pragma-inserted C/C++ source files as the code input. We concatenate the design files into a single token sequence and feed it to the frozen code encoder. We use only design code because the kernel baseline $y_k$ is predicted from the kernel IR graph alone, while code embeddings serve as an auxiliary signal for the design delta. Figure~\ref{fig:code_examples} shows an example kernel and its corresponding pragma-inserted design.

\subsection{Differential Heads}
To implement the decomposition in Eq.~\eqref{eq:delta_def}, we use two prediction heads that are trained jointly but serve distinct roles. The detailed architecture is illustrated in Fig.~\ref{fig:overview}. The kernel head estimates the kernel baseline solely from the kernel embedding:
\begin{equation}
    \hat{y}_k = f_k(\mathbf{h}_k),
\end{equation}
where $f_k:\mathbb{R}^{D}\!\to\!\mathbb{R}$ is a small MLP regressor. Concretely, both heads use the same MLP template, differing only in input dimension.

The delta head targets the design-induced change and therefore conditions on both the kernel and design embeddings, optionally augmented with the code embedding $\mathbf{h}_c$, as shown in Algorithm~\ref{alg:diffhls}:
\begin{equation}
    \mathbf{h}_{\Delta} =
    \begin{cases}
    [\mathbf{h}_k;\mathbf{h}_d], & \text{w/o code emb.} \\
    [\mathbf{h}_k;\mathbf{h}_d;\mathbf{h}_c], & \text{w/ code emb.}
    \end{cases}
\end{equation}
\begin{equation}
    \widehat{\Delta} = f_{\Delta}(\mathbf{h}_{\Delta}),
\end{equation}
where $f_{\Delta}:\mathbb{R}^{2D}\!\to\!\mathbb{R} \text{~or } \mathbb{R}^{3D}\!\to\!\mathbb{R}$ is another MLP.

We then compose the final design prediction as:
\begin{equation}
    \hat{y}_d = \hat{y}_k + \widehat{\Delta}.
\end{equation}
This two-head design separates the kernel-dependent baseline from pragma-driven changes. Code embeddings enter only through the delta pathway, so $\hat{y}_k$ remains a graph-only kernel baseline.

\subsection{Training Objective}
We supervise kernel, delta, and design outputs with a SmoothL1 loss $\ell(\cdot,\cdot)$:
\begin{equation}
    \mathcal{L} = \ell(\hat{y}_k, y_k) + \ell(\widehat{\Delta}, \Delta) + \ell(\hat{y}_k+\widehat{\Delta}, y_d),
\label{eq:train_obj}
\end{equation}
where all three terms are equally weighted. The first two terms directly supervise the kernel baseline prediction and the delta prediction, respectively. The third term enforces consistency between the composed prediction $\hat{y}_k+\widehat{\Delta}$ and the ground-truth design target $y_d$.

This three-term objective serves two purposes. First, it ensures that both the kernel head and delta head learn meaningful predictions individually. Second, it prevents error accumulation when composing the two predictions. Without the consistency term, errors in $\hat{y}_k$ and $\widehat{\Delta}$ could accumulate and produce larger errors in the final design prediction. The consistency term $\ell(\hat{y}_k+\widehat{\Delta}, y_d)$ penalizes such misalignment and encourages the two heads to cooperate.

\begin{table}[!t]
\centering
    \caption{Target value ranges (kernel, delta, design).}
    \vspace{-1em}
    \label{tab:dataset_stats}
        \begin{tabular*}{\columnwidth}{@{\extracolsep{\fill}}c|ccc}
        \toprule
        \textbf{[min, max]} & \textbf{Kernel} & \textbf{Delta} & \textbf{Design} \\
        \midrule
        DSP & [3, 29] & [$-$7, 75] & [5, 94] \\
        FF  & [495, 3994] & [$-$174, 25414] & [1296, 28967] \\
        LUT & [1214, 5192] & [151, 35865] & [1842, 39464] \\
        CP  & [5.31, 7.21] & [$-$0.10, 0.78] & [5.55, 7.75] \\
        \bottomrule
        \end{tabular*}%
\vspace{-1em}
\end{table}

\section{Experimental Setup}

\subsection{Metrics}
We report mean absolute error, denoted MAE; mean absolute percentage error, denoted MAPE and reported in percent; and coefficient of determination, denoted $R^2$. Lower is better for MAE and MAPE, and higher is better for $R^2$.
For a set of $N$ designs with targets $\{y_i\}$ and predictions $\{\hat{y}_i\}$, MAE, MAPE, and $R^2$ are:
\begin{equation}
\label{eq:metrics}
\begin{aligned}
\mathrm{MAE} &= \frac{1}{N}\sum_{i=1}^N |y_i-\hat{y}_i|, \\
\mathrm{MAPE} &= \frac{100}{N_0}\sum_{i:\, y_i\neq 0} \left|\frac{y_i-\hat{y}_i}{y_i}\right|, \\
R^2 &= 1 - \frac{\sum_{i=1}^N (y_i-\hat{y}_i)^2}{\sum_{i=1}^N (y_i-\bar{y})^2}.
\end{aligned}
\end{equation}
Here $N_0=\bigl|\{i:\, y_i\neq 0\}\bigr|$, $\bar{y}=\frac{1}{N}\sum_{i=1}^N y_i$, and MAPE excludes samples with zero targets to avoid division-by-zero.

\begin{table}[!t]
    \centering
            \caption{QoR prediction accuracy under different GNN backbones.}
            \vspace{-1em}
            \label{tab:resource_prediction}
            \resizebox{\columnwidth}{!}{%
            \begin{threeparttable}
                \begin{tabular}{c|l|ccccc}
    \toprule
                \textbf{Backbone}
                & \textbf{Method}
                & \textbf{DSP} & \textbf{FF} & \textbf{LUT} & \textbf{CP} \\
    \midrule
                \multirow{3}{*}{\centering PNA}
                & GNN~\cite{wu2022high} & 14.71 & 26.47 & 22.86 & 8.87 \\
                & HGNN~\cite{gao2024hierarchical}     &  7.84 &  9.65 & 10.55 & - \\
                & \DiffHLS                & \textbf{3.63} &  \textbf{8.18} &  \textbf{5.52} &  \textbf{1.34} \\
    \midrule
                \multirow{3}{*}{\centering GraphSAGE}
                & GNN~\cite{wu2022high} & 17.01 & 39.11 & 28.09 & 8.25 \\
                & HGNN~\cite{gao2024hierarchical}     &  6.94 &  9.99 &  9.86 & - \\
                & \DiffHLS                & \textbf{3.31} &  \textbf{7.34} &  \textbf{5.12} &  \textbf{1.06} \\
    \midrule
                \multirow{3}{*}{\centering GCN}
                & GNN~\cite{wu2022high} & 25.30 & 38.34 & 28.64 & 8.79 \\
                & HGNN~\cite{gao2024hierarchical}     &  \textbf{7.45} &  11.24 &  11.27 & - \\
                & \DiffHLS                & 8.10 &  \textbf{9.31} &  \textbf{5.46} &  \textbf{1.37} \\
    \midrule
                \multirow{3}{*}{\centering GAT}
                & GNN~\cite{wu2022high} & 28.66 & 54.73 & 46.19 & 10.32 \\
                & HGNN~\cite{gao2024hierarchical}     &  8.18 &  11.57 &  12.34 & - \\
                & \DiffHLS                & \textbf{6.41} &  \textbf{9.63} &  \textbf{6.11} &  \textbf{1.53} \\
    \bottomrule
                \end{tabular}%
                \begin{tablenotes}[flushleft]
                \footnotesize
                \item We report MAPE in \%. Best results are highlighted in \textbf{bold}.
                \end{tablenotes}
            \end{threeparttable}
            }
\vspace{-1em}
\end{table}

\begin{table*}[!t]
    \centering
        \caption{Ablation study: differential vs. direct prediction and effect of code embeddings.}
        \vspace{-1em}
        \label{tab:ablation_method_variants}
        \begin{threeparttable}
            \resizebox{\textwidth}{!}{%
            \begin{tabular}{c|l|ccc|ccc|ccc|ccc}
    \toprule
            \multirow{2}{*}{\textbf{Backbone}}
            & \multirow{2}{*}{\textbf{Method}}
            & \multicolumn{3}{c|}{\textbf{DSP}}
            & \multicolumn{3}{c|}{\textbf{FF}}
            & \multicolumn{3}{c|}{\textbf{LUT}}
            & \multicolumn{3}{c}{\textbf{CP}} \\
            &
            & MAPE~$\downarrow$ & MAE~$\downarrow$ & $R^2$~$\uparrow$
            & MAPE~$\downarrow$ & MAE~$\downarrow$ & $R^2$~$\uparrow$
            & MAPE~$\downarrow$ & MAE~$\downarrow$ & $R^2$~$\uparrow$
            & MAPE~$\downarrow$ & MAE~$\downarrow$ & $R^2$~$\uparrow$
            \\
    \midrule
            \multirow{3}{*}{\centering PNA}
            & \DiffHLS~(w/o diff)  & 3.86 & \textbf{1.16} & \textbf{0.980} & 10.75 & 725.94 & 0.972 & 8.50 & 891.66 & 0.969 & \textbf{1.22} & \textbf{0.090} & 0.711 \\
            & \DiffHLS~(w/o code emb.)   & 4.18 & 1.39 & 0.976 & 8.51 & 677.34 & 0.963 & 6.30 & 641.35 & 0.967 & \textbf{1.22} & \textbf{0.090} & \textbf{0.735} \\
            & \DiffHLS             & \textbf{3.63} & 1.21 & 0.899 & \textbf{8.18} & \textbf{629.84} & \textbf{0.973} & \textbf{5.52} & \textbf{627.03} & \textbf{0.989} & 1.34 & 0.099 & 0.710 \\
            \midrule
            \multirow{3}{*}{\centering GraphSAGE}
            & \DiffHLS~(w/o diff)  & 4.19 & \textbf{1.25} & \textbf{0.971} & 10.60 & 680.01 & 0.926 & 7.76 & 707.05 & 0.935 & 1.31 & 0.096 & 0.679 \\
            & \DiffHLS~(w/o code emb.)   & 5.11 & 1.81 & 0.945 & 8.64 & 698.86 & \textbf{0.974} & 5.78 & 576.78 & \textbf{0.991} & 1.24 & 0.092 & 0.721 \\
            & \DiffHLS             & \textbf{3.31} & 1.28 & 0.964 & \textbf{7.34} & \textbf{647.85} & 0.897 & \textbf{5.12} & \textbf{510.83} & \textbf{0.991} & \textbf{1.06} & \textbf{0.078} & \textbf{0.847} \\
            \midrule
            \multirow{3}{*}{\centering GCN}
            & \DiffHLS~(w/o diff)  & 8.91 & \textbf{2.78} & 0.937 & 10.71 & \textbf{897.59} & \textbf{0.935} & 7.83 & 776.59 & \textbf{0.986} & \textbf{1.22} & \textbf{0.090} & 0.718 \\
            & \DiffHLS~(w/o code emb.)   & 10.08 & 3.34 & 0.915 & 10.37 & 996.00 & 0.678 & 6.68 & 751.33 & 0.981 & 1.27 & 0.094 & \textbf{0.732} \\
            & \DiffHLS             & \textbf{8.10} & 2.84 & \textbf{0.950} & \textbf{9.31} & 1001.49 & 0.413 & \textbf{5.46} & \textbf{621.47} & 0.985 & 1.37 & 0.101 & 0.664 \\
            \midrule
            \multirow{3}{*}{\centering GAT}
            & \DiffHLS~(w/o diff)  & 7.67 & 2.57 & 0.938 & 10.88 & 820.99 & 0.947 & 13.39 & 1186.50 & 0.969 & 1.32 & 0.098 & 0.294 \\
            & \DiffHLS~(w/o code emb.)   & 8.07 & 2.60 & 0.952 & 9.98 & \textbf{763.87} & \textbf{0.967} & \textbf{5.88} & \textbf{562.37} & \textbf{0.991} & \textbf{1.28} & \textbf{0.095} & \textbf{0.700} \\
            & \DiffHLS             & \textbf{6.41} & \textbf{2.18} & \textbf{0.958} & \textbf{9.63} & 905.99 & 0.863 & 6.11 & 715.22 & 0.982 & 1.53 & 0.113 & 0.628 \\
    \bottomrule
            \end{tabular}%
            }
            \begin{tablenotes}[flushleft]
            \footnotesize
            \item \DiffHLS~(w/o diff) disables differential learning and directly predicts design targets without the kernel--design delta decomposition.
            \item \DiffHLS~(w/o code emb.) disables code embeddings.
            \end{tablenotes}
        \end{threeparttable}
\vspace{-1em}
    \end{table*}

\subsection{Dataset}
We evaluate on 27 PolyBench/C kernels~\cite{Polybench} and 10{,}108 design points derived from the ForgeHLS PolyBench subset~\cite{peng_forgehls_2025} via pragma insertion. We predict design-level QoR targets $y_d \in \{\text{DSP},\text{FF},\text{LUT},\text{CP}\}$, whose value ranges are listed in Table~\ref{tab:dataset_stats}.
We use a design-level random split with an 8:1:1 train/validation/test ratio over paired kernel--design samples.
We train separate models for each of DSP, FF, LUT, and CP.

\subsection{Implementation Details}
Unless otherwise stated, we train with Adam optimizer, learning rate $5\times 10^{-4}$, batch size 16, hidden size 128, two GNN layers, and dropout 0.02. We use ReduceLROnPlateau with patience 15 and factor 0.8, and select the best checkpoint based on validation loss. All experiments are run on a single NVIDIA RTX 4090 GPU with 24\,GB memory.

\begin{table*}[!t]
    \centering
    \caption{Ablation study: prediction head accuracy.}
    \vspace{-1em}
    \label{tab:ablation_head_accuracy}
    \begin{threeparttable}
        \resizebox{\textwidth}{!}{%
        \begin{tabular}{c|c|c|ccc|ccc|ccc|ccc}
        \toprule
        \multirow{2}{*}{\textbf{Head}}
        & \multirow{2}{*}{\textbf{Backbone}}
        & \multirow{2}{*}{\shortstack{\textbf{Code}\\\textbf{Emb.}}}
        & \multicolumn{3}{c|}{\textbf{DSP}}
        & \multicolumn{3}{c|}{\textbf{FF}}
        & \multicolumn{3}{c|}{\textbf{LUT}}
        & \multicolumn{3}{c}{\textbf{CP}} \\
        & & &
        MAPE & MAE & $R^2$ &
        MAPE & MAE & $R^2$ &
        MAPE & MAE & $R^2$ &
        MAPE & MAE & $R^2$
        \\
        \midrule
        \multirow{8}{*}{Kernel $\hat{y}_k$}
        & \multirow{2}{*}{PNA}
        & No  & \textbf{0.43} & \textbf{0.08} & \textbf{1.000} & 0.74 & 15.53 & 0.999 & 0.55 & 13.36 & 1.000 & \textbf{0.06} & \textbf{0.004} & \textbf{0.993} \\
        &     & Yes & 0.84 & 0.12 & 0.999 & \textbf{0.66} & \textbf{15.08} & \textbf{1.000} & \textbf{0.38} & \textbf{8.81} & \textbf{1.000} & 0.12 & 0.008 & 0.933 \\
        \cmidrule(lr){2-15}
        & \multirow{2}{*}{GraphSAGE}
        & No  & 0.89 & 0.14 & 0.999 & 0.94 & \textbf{20.13} & 1.000 & \textbf{0.58} & \textbf{13.54} & 1.000 & 0.11 & 0.008 & 0.955 \\
        &     & Yes & \textbf{0.51} & \textbf{0.09} & \textbf{1.000} & \textbf{0.86} & 20.48 & 1.000 & 0.61 & 13.83 & 1.000 & \textbf{0.07} & \textbf{0.005} & \textbf{0.998} \\
        \cmidrule(lr){2-15}
        & \multirow{2}{*}{GCN}
        & No  & 1.20 & 0.15 & \textbf{0.999} & \textbf{0.67} & \textbf{13.44} & 1.000 & 0.82 & 19.48 & 0.999 & \textbf{0.06} & \textbf{0.004} & \textbf{0.996} \\
        &     & Yes & \textbf{0.86} & \textbf{0.14} & 0.999 & 0.74 & 16.36 & \textbf{1.000} & \textbf{0.38} & \textbf{9.05} & \textbf{1.000} & 0.24 & 0.017 & 0.379 \\
        \cmidrule(lr){2-15}
        & \multirow{2}{*}{GAT}
        & No  & \textbf{0.79} & \textbf{0.12} & \textbf{0.999} & 2.09 & 51.36 & 0.994 & \textbf{0.91} & \textbf{22.56} & \textbf{0.999} & \textbf{0.10} & \textbf{0.007} & \textbf{0.992} \\
        &     & Yes & 1.63 & 0.25 & 0.997 & \textbf{1.01} & \textbf{21.56} & \textbf{0.999} & 1.16 & 30.21 & 0.998 & 0.12 & 0.008 & 0.971 \\
        \midrule
        \multirow{8}{*}{Delta $\widehat{\Delta}$}
        & \multirow{2}{*}{PNA}
        & No  & 13.55 & 1.38 & \textbf{0.971} & \textbf{58.52} & 676.20 & 0.961 & 13.71 & 641.63 & 0.965 & \textbf{35.80} & \textbf{0.090} & \textbf{0.709} \\
        &     & Yes & \textbf{11.37} & \textbf{1.23} & 0.872 & 66.76 & \textbf{631.17} & \textbf{0.972} & \textbf{12.07} & \textbf{627.14} & \textbf{0.988} & 38.13 & 0.099 & 0.654 \\
        \cmidrule(lr){2-15}
        & \multirow{2}{*}{GraphSAGE}
        & No  & 18.59 & 1.81 & 0.931 & 53.23 & 695.03 & \textbf{0.973} & 11.83 & 576.72 & 0.990 & 36.61 & 0.092 & 0.706 \\
        &     & Yes & \textbf{11.92} & \textbf{1.28} & \textbf{0.956} & \textbf{46.15} & \textbf{650.10} & 0.891 & \textbf{10.63} & \textbf{511.54} & \textbf{0.991} & \textbf{31.33} & \textbf{0.078} & \textbf{0.740} \\
        \cmidrule(lr){2-15}
        & \multirow{2}{*}{GCN}
        & No  & 38.82 & 3.37 & 0.895 & 59.39 & \textbf{995.09} & \textbf{0.661} & 13.60 & 754.88 & 0.980 & \textbf{36.13} & \textbf{0.093} & \textbf{0.675} \\
        &     & Yes & \textbf{26.43} & \textbf{2.86} & \textbf{0.938} & \textbf{32.41} & 1002.12 & 0.384 & \textbf{11.72} & \textbf{622.58} & \textbf{0.984} & 37.80 & 0.102 & 0.638 \\
        \cmidrule(lr){2-15}
        & \multirow{2}{*}{GAT}
        & No  & 29.27 & 2.59 & 0.939 & 125.26 & \textbf{754.27} & \textbf{0.966} & \textbf{13.14} & \textbf{563.01} & \textbf{0.991} & \textbf{33.77} & \textbf{0.094} & \textbf{0.657} \\
        &     & Yes & \textbf{22.45} & \textbf{2.17} & \textbf{0.948} & \textbf{59.00} & 906.03 & 0.857 & 12.63 & 713.02 & 0.981 & 39.63 & 0.112 & 0.585 \\
        \bottomrule
        \end{tabular}%
        }
    \end{threeparttable}
\vspace{-1em}
\end{table*}

\section{Experimental Results}
\subsection{Main Results}
Table~\ref{tab:resource_prediction} summarizes design-level QoR prediction MAPE under four GNN backbones. \DiffHLS~achieves strong performance across all backbones, with GraphSAGE yielding the best results. Compared to the GNN baseline~\cite{wu2022high}, \DiffHLS~reduces MAPE substantially on every target under all backbones. For example, under GraphSAGE, DSP MAPE drops from 17.01\% to 3.31\% and LUT MAPE drops from 28.09\% to 5.12\%.

\begin{figure}[!t]
    \centering
    \includegraphics[width=0.9\columnwidth]{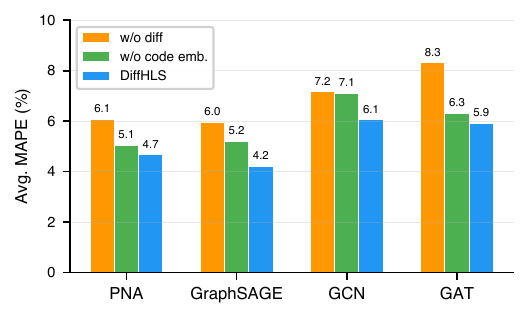}
    \vspace{-1em}
    \caption{Average MAPE (\%) across DSP, FF, LUT, and CP for the three method variants under four GNN backbones.}
    \label{fig:ablation_bar}
    \vspace{-1em}
\end{figure}

\subsection{Differential vs. Direct Prediction}
Table~\ref{tab:ablation_method_variants} compares \DiffHLS~with a direct prediction variant (\DiffHLS~w/o diff) that disables the differential formulation and directly predicts design targets. Disabling differential learning consistently increases MAPE on FF and LUT across all backbones. The degradation is particularly large on LUT under GAT, where MAPE increases from 6.11\% to 13.39\%. 

This validates our core hypothesis: by decomposing $y_d = y_k + \Delta$, the model only needs to learn the pragma-induced change $\Delta$ rather than the full target value, which varies by orders of magnitude across different kernels. The differential formulation therefore simplifies the learning problem considerably. Fig.~\ref{fig:ablation_bar} visualizes the average MAPE of the three variants across all backbones.

\subsection{Effect of Code Embeddings}
Table~\ref{tab:ablation_method_variants} also compares \DiffHLS~with a GNN-only variant (\DiffHLS~w/o code emb.) that removes code embeddings from the delta pathway. Adding code embeddings consistently decreases MAPE across DSP, FF, and LUT under all backbones. The largest impact is on DSP, where MAPE decreases from 5.11\% to 3.31\% under GraphSAGE. 

This suggests that LLM code embeddings capture pragma annotations and loop structures that may be lost during IR generation, providing complementary cues to the graph-based features.


\subsection{Prediction Head Accuracy}
Table~\ref{tab:ablation_head_accuracy} reports accuracy for the kernel head and delta head separately. The kernel head achieves MAPE below 2\% and $R^2$ above 0.99 on DSP, FF, and LUT, confirming that kernel baselines are well-captured by IR graph structure alone. The delta head is more challenging due to the wide range of pragma-induced QoR changes, but still maintains reasonable $R^2$. Adding code embeddings reduces delta MAPE on DSP from 13.55\% to 11.37\% under PNA and from 18.59\% to 11.92\% under GraphSAGE. These results confirm that the kernel baseline is an easy subproblem, while the delta head handles the harder pragma-driven variation.

\begin{table*}[!t]
    \centering
                \caption{Ablation study: code LLM backbone.}
                \vspace{-1em}
                \label{tab:ablation_llm_backbone}
                \begin{threeparttable}
                    \resizebox{\textwidth}{!}{%
                    \begin{tabular}{c|l|ccc|ccc|ccc|ccc}
    \toprule
                    \multirow{2}{*}{\textbf{Backbone}}
                    & \multirow{2}{*}{\textbf{Model}}
                    & \multicolumn{3}{c|}{\textbf{DSP}}
                    & \multicolumn{3}{c|}{\textbf{FF}}
                    & \multicolumn{3}{c|}{\textbf{LUT}}
                    & \multicolumn{3}{c}{\textbf{CP}} \\
                    & & MAPE~$\downarrow$ & MAE~$\downarrow$ & $R^2$~$\uparrow$
                    & MAPE~$\downarrow$ & MAE~$\downarrow$ & $R^2$~$\uparrow$
                    & MAPE~$\downarrow$ & MAE~$\downarrow$ & $R^2$~$\uparrow$
                    & MAPE~$\downarrow$ & MAE~$\downarrow$ & $R^2$~$\uparrow$
                    \\
                    \midrule
            \multirow{2}{*}{\centering PNA}
                    & Qwen2.5-Coder-1.5B & \textbf{3.63} & \textbf{1.21} & 0.899 & 8.18 & 629.84 & \textbf{0.973} & \textbf{5.52} & 627.03 & \textbf{0.989} & 1.34 & 0.099 & 0.710 \\
                    & Llama-3.2-1B & 3.96 & 1.40 & \textbf{0.979} & \textbf{8.09} & \textbf{524.57} & 0.967 & 5.60 & \textbf{457.07} & 0.984 & \textbf{1.13} & \textbf{0.084} & \textbf{0.776} \\
    \midrule
            \multirow{2}{*}{\centering GraphSAGE}
                    & Qwen2.5-Coder-1.5B & \textbf{3.31} & \textbf{1.28} & \textbf{0.964} & \textbf{7.34} & 647.85 & 0.897 & \textbf{5.12} & 510.83 & \textbf{0.991} & \textbf{1.06} & \textbf{0.078} & \textbf{0.847} \\
                    & Llama-3.2-1B & 4.52 & 1.71 & 0.926 & 8.14 & \textbf{528.27} & \textbf{0.963} & 5.56 & \textbf{453.39} & 0.989 & 1.17 & 0.086 & 0.738 \\
    \midrule
            \multirow{2}{*}{\centering GCN}
                    & Qwen2.5-Coder-1.5B & 8.10 & 2.84 & \textbf{0.950} & 9.31 & 1001.49 & 0.413 & \textbf{5.46} & \textbf{621.47} & 0.985 & 1.37 & 0.101 & 0.664 \\
                    & Llama-3.2-1B & \textbf{8.08} & \textbf{2.73} & 0.947 & \textbf{8.93} & \textbf{817.44} & \textbf{0.947} & 6.50 & 717.54 & \textbf{0.988} & \textbf{1.13} & \textbf{0.083} & \textbf{0.774} \\
    \midrule
            \multirow{2}{*}{\centering GAT}
                    & Qwen2.5-Coder-1.5B & 6.41 & 2.18 & \textbf{0.958} & 9.63 & 905.99 & 0.863 & 6.11 & 715.22 & 0.982 & 1.53 & 0.113 & \textbf{0.628} \\
                    & Llama-3.2-1B & \textbf{5.91} & \textbf{2.09} & 0.957 & \textbf{9.31} & \textbf{685.07} & \textbf{0.958} & \textbf{6.06} & \textbf{602.63} & \textbf{0.991} & \textbf{1.20} & \textbf{0.088} & 0.628 \\
    \bottomrule
                    \end{tabular}%
                    }
                \end{threeparttable}
    \vspace{-1em}
\end{table*}

\begin{table}[!t]
    \centering
        \caption{ForgeHLS large-scale evaluation.}
        \vspace{-1em}
        \label{tab:forgehls_large}
        \begin{threeparttable}
            \begin{tabular}{c|cccc}
            \toprule
            \textbf{Backbone} & \textbf{DSP} & \textbf{FF} & \textbf{LUT} & \textbf{CP} \\
            \midrule
            PNA & 5.70 & 25.75 & 10.05 & \textbf{4.21} \\
            GraphSAGE & \textbf{5.69} & \textbf{17.30} & \textbf{7.77} & 4.93 \\
            GCN & 9.11 & 25.18 & 11.94 & 5.83 \\
            GAT & 10.75 & 20.16 & 9.27 & 4.41 \\
            \bottomrule
            \end{tabular}%
            \begin{tablenotes}[flushleft]
            \footnotesize
            \item We report design-level MAPE in \%.
            \end{tablenotes}
        \end{threeparttable}
    \vspace{-1em}
\end{table}

\subsection{GNN Backbone}
Across all experiments, GraphSAGE consistently yields the best results, outperforming PNA, GCN, and GAT on all QoR targets. The differential formulation benefits all four backbones substantially, and code embeddings provide further gains under GraphSAGE. Overall, \DiffHLS~is robust to the choice of GNN backbone.

\subsection{Code LLM Backbone}
Table~\ref{tab:ablation_llm_backbone} compares two code LLM backbones for embedding extraction under the same training protocol. Under PNA and GraphSAGE, Qwen2.5-Coder-1.5B achieves lower MAPE on DSP and LUT, while Llama-3.2-1B sometimes yields better MAE and $R^2$ on FF. Under GCN, Llama-3.2-1B achieves lower MAPE on DSP, FF, and CP, while Qwen2.5-Coder-1.5B retains an advantage on LUT. Under GAT, Llama-3.2-1B outperforms Qwen2.5-Coder-1.5B on most metrics. These results suggest that the optimal code backbone may depend on the GNN architecture and target metric. Overall, we use Qwen2.5-Coder-1.5B as the default code embedding backbone.

\subsection{ForgeHLS Large-Scale Evaluation}
To further validate scalability, we evaluate \DiffHLS~on the ForgeHLS large-scale dataset~\cite{peng_forgehls_2025}. All runs use the full \DiffHLS~configuration with differential heads and code embeddings enabled.

Table~\ref{tab:forgehls_large} reports design-level MAPE for four GNN backbones. GraphSAGE delivers the strongest overall performance, achieving the lowest MAPE on resource targets. PNA achieves the lowest CP MAPE. The ranking among backbones is consistent with the PolyBench results, confirming that \DiffHLS~generalizes to a substantially larger and more diverse design space.

\section{Conclusion}
We presented \DiffHLS, a differential learning framework for HLS QoR prediction that learns from kernel--design pairs by decomposing design-level targets into a kernel baseline and a design-induced delta, and augments graph representations with LLM code embeddings. On PolyBench, \DiffHLS~improves design-level prediction accuracy over baselines under four GNN backbones across all four QoR targets. Our ablations show that differential learning provides the largest gain, and code embeddings further improve performance. We also validate scalability on the larger ForgeHLS dataset.

\bibliographystyle{IEEEtran}
\bibliography{references}

\end{document}